# A Deep Learning Technique using a Sequence of Follow Up X-Rays for Disease classification

**Sairamvinay Vijayaraghavan**
University of California, Davis

**David Haddad**
University of California, Davis

**Shikun Huang**
University of California, Davis

**Seongwoo Choi**
University of California, Davis

## Abstract

The ability to predict lung and heart based diseases using deep learning techniques is central to many researchers, particularly in the medical field around the world. In this paper, we present a unique outlook of a very familiar problem of disease classification using X-rays. We present a hypothesis that X-rays of patients included with their followup history of their most recent three chest X-ray images would perform better in disease classification in comparison to one chest X-ray image input using an internal CNN to perform feature extraction. We have discovered that our generic deep learning architecture which we propose for solving this problem performs well with 3 input X-ray images provided per sample for each patient. In this paper, we have also established that without additional layers before the output classification, the CNN models will improve the performance of predicting the disease labels for each patient. We have provided our results in ROC curves and AUROC scores. We define a fresh approach of collecting three X-ray images for training deep learning models, which we have concluded that it has clearly improved the performance of the models. We have shown that ResNet, in general has a better result than any other CNN model used in the feature extraction phase. With our original approach on data pre-processing, image training, and pre-trained models, we believe that the current research will assist many medical institutions around the world, and this will improve the prediction of patients' symptoms and diagnose them with more accurate cure.

Note: Our project source code can be found at: `https://github.com/Sairamvinay/Chest-X-ray-DL`

## 1 Introduction

Deep Learning in Computer Vision has found its importance in medicinal AI. Computer Vision techniques and deep learning based predictive models go hand-in-hand in case of various subsections in healthcare which involve different image based data: such as X-rays, PT scans. There has been a lot of work in predicting diseases for patients using X-rays using deep learning models such as Convolutional Neural Networks [15][10][1][9], Support Vector Machines and K- Nearest Neighbors [5][16]. However, most of these models use strictly one X-ray image per sample for training the model for predicting disease labels as a multi classification problem. Our team decided to present a slightly unique setting for a familiar problem. In this paper, we are going to present a deep learning model, which takes in a sequence of the consecutive previous chest X-rays of a patient (termed shortly as follow-ups) and analyzes the variation and difference across this sequence. For the feature extraction phase of the images, the model would involve convolutional neural networks (CNN). We have another phase in our model which would involve an optional Long Short Term Memory layer for capturing the time based temporal relationship across the pixels three images. The goal is to distinguish between the fourteen different respiratory diseases and also the No Finding label (indicating normal) using the previous three followup X-rays of every patient by predicting the disease label associated with the third X-ray image. We frame it as a multi-classification problem by predicting any of the 15 disease labels (which includes the 'No Finding' as a label). For feature extraction, we plan to use three different existent deep CNN models such as DenseNet[4], MobileNet[12] and ResNet[3] as the CNNs for the X-ray image inputs. For all our experiments, we had used the large X-ray dataset from NIH which was openly available on Kaggle and we had used the entire dataset consisting of 112K X-rays for around 30K patients[15].

Through our research, we have discovered that there is a total of fourteen different diseases that are related to our breathing system and we could observe the normal state of the body. In order to analyze normal and disease based X-ray samples, we have decided to keep the "No Finding" label. We would also plan to analyze the PA (Posterior Anterior) view X-rays and AP (Anterior Posterior) view X-rays separately due to the impact that the X-ray View positions might have on the disease labels. It is important to inform the difference between the PA and AP views of chest X-rays. Posterior-Anterior (PA) projection is the standard projection with higher quality and accuracy to assess heart size than AP images, but PA projection is not always possible. PA view position is produced by X-ray traversing the patient from posterior to anterior and striking the film. Anterior-Posterior (AP) projection is used the most when it is not possible for radiographers to acquire a PA chest X-ray, and AP view position is produced with the X-ray traversing the patient from anterior to posterior striking the film[8]. More of this work is discussed in the methodology section under the datasets subsection.

In addition to the above experiments, we also plan to compare and analyze specifically the impact of LSTMs on these X-ray based inputs. So, we had trained two types of models: with LSTM and without LSTM layer in our architecture (more on this is in subsection "Basic Algorithm for the Modeling").

Throughout this paper, we intend to present a single deep learning framework which would take in more than one X-ray per patient for analysis and would intend to treat these X-rays as an image sequence which would be then used for predicting the disease label based on the differences observed within the regions present across each follow up X-ray. Our goal to identify how does followup X-ray images play a significant role in predicting the disease labels. We also felt that the difference across the 3 X-ray images can be captured by deep learning CNN models which would play a huge role in the disease. Here's some examples which show how an X-ray change over time.

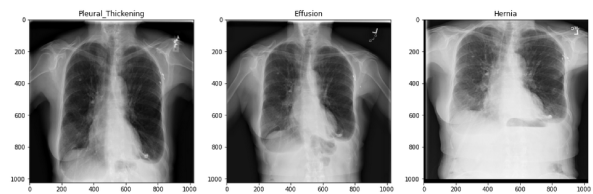

Figure 1: 3 followup X-ray images of a patient in PA view with the 3rd followup labelled as Hernia

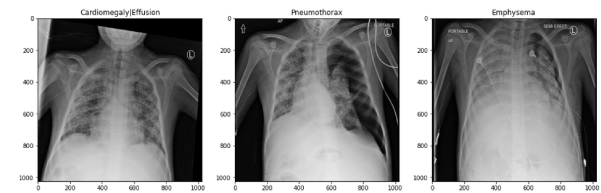

Figure 2: 3 followup X-ray images of a patient in AP view with the 3rd followup labelled as Emphysema

Our team was strongly influenced and motivated by previous research on analyzing Chest X-rays to distinguish symptoms and predict the disease associated with the patient. We emphasize that most of the previous research focused on training on single Chest X-rays for disease classification. To our best knowledge, we are presenting a unique setting of a very familiar problem. Previous research mostly involved machine learning based multi classification of images without any particular pre-filtering procedure of samples which is based on the number of follow-ups per every patient. Rather works were conducted with classifying each of the disease labels diagnosed with for every patient regardless of the followup associated in the dataset.

## 2 Related Work

There have been several approaches to analyze chest X-rays using deep neural networks. One of the prior works does explore the classification of the Chest X-rays using ResNet and DenseNet for feature extraction for X-ray images [1]. This paper uses all of the X-ray images for disease classification without even considering followups or view position associated with the X-ray. Our approach distinguishes from those researchers that we have been analyzing the datasets and we wanted to see a closer look into datasets and wanted to know if three follow-ups could boost up the performance of the neural network we were building.

We could discover that our approach was derived mainly from the 'ChestNet' paper [14]. This paper is probably the closest research paper that is similar to the project we are working on. The authors explore the probability of each possible disease of the patient using sigmoid output activation. They had used a seq2seq model with attention for fine tuning on the convolved output of hte X-ray after feature extraction X-rays and they had also used GradCAM technique [13] for identifying key regions in the X-ray image. However, they did not explain the impact of training models with CNNs followed by LSTMs and how they could change the overall performance of the classification model. We also explored how if we do not use LSTM would change the performance of the neural network classification.

There are also works which have primarily used PCA based feature reduction and simple statistical-based feature selection techniques such as forward-backward selection techniques can help in X-ray images based feature extraction. They also trained support vector machine (SVM) and K-nearest neighbors (KNN) were applied to classify the esophageal cancer images with respect to their specific types. They had performed this task for a specific task of esophageal cancer [16]. In addition to the above work, texture and edge features on a local and global level within an X-ray image are extracted while using KNNs and SVMs again for disease classification [5]. Similar to this feature extraction technique, there is another work where the authors describe how feature extraction on TB images is important for disease analysis by using PCA and kPCA and training a linear regression classification model to predict the disease [11].

CheXNeXt model was a deep learning model from which we had been heavily motivated to formulate our hypothesis. CheXNeXt is a convolutional neural network that concurrently detects the presence of 14 pathologies [9].

Some of the researchers also have used DenseNet based CNN architectures for X-ray based classification[1]. CheXNet is a 121-layer Dense Convolutional Network based on DenseNet is proposed to perform better than practicing radiologists at pneumonia detection on frontal view X-rays [10]. The model is trained on the same dataset that we are also using for this paper and the results show an increase in AUROC scores over the state of the art algorithms. Research also shows that CNN's and transfer learning can improve detection of pneumonia and COVID patients[7]. This paper utilizes Grad-CAM to illustrate the radiology disease detection for all their experiments.

## 3 Significance of this project

In healthcare, Chest X-rays are expensive and are very frequently used diagnostic examinations. Chest X-rays can play a significant role in determining the lung/heart diseases associated[2]. Practically, doctors intend to use the history of a patient's health vitals and medical results for analyzing his health conditions. For example, doctors would find it very useful to analyze the current as well as the previous visit X-rays for analyzing diseases in the long term for a patient. Temporal analysis of X-ray history is very critical and it would facilitate the work of a medical doctor in analyzing such work using robust deep learning models.

While there has been many works done on disease classification using single X-ray images[15][2][14], a doctor would refer to prior X-ray history for obtaining a much more accurate picture of the changes occurring within the chest/lungs and hence would be able to analyze this problem much better. To our best knowledge, we present a paper which instead of taking one image per sample for classification, we present a model which treats the X-rays by itself as a image sequence which would be studied as a temporal sequence of inputs.

We intend to mimic what a medical doctor would intend to do practically by analyzing the previous few X-rays at once and then predict the current disease label for a patient. For our paper, we wanted to focus only on X-rays. The intention of this paper will lead to an easier approach for the medical institutions in predicting the chest/lung based diseases using previous three X-rays. This will result in facilitating the medical doctor in analyzing 3 previous X-rays for disease classification without having to observe it visually and can rather use this model for finer results.

In order to validate this significance we propose through this paper, we plan to conduct a short experiment using single X-ray image based disease classification. We had used a pre-trained DenseNet121 model on all the X-ray images (112K) for performing image based disease classification for just a single X-ray. We had used DenseNet121 pre-trained model [6] and we had used it to evaluate on the single X-rays dataset split up as either AP or PA view position. We had performed Receiver Operator Characteristic Area Under Curve (ROC AUC) score analysis. For this task, we had not used the No-Finding label and we had used a pre-trained DenseNet121 which again had a sigmoid activation[6]. So, we had compared this with our best DenseNet model for the three X-ray image based task. More on these results in Results section.

## 4 Methodology

### 4.1 Overall Pipeline

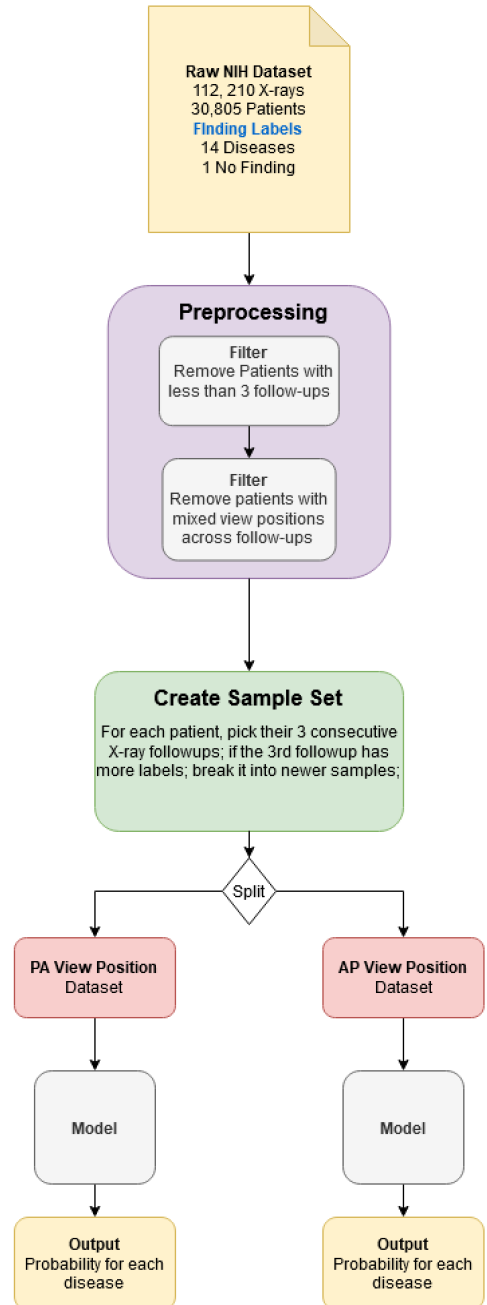

Figure 3: Pipeline describing the project proceedure

### 4.2 Dataset

NIH provides a file named "Dataentry2017.csv" containing patient data such as Image Index (indicating the image filename), Finding Labels (the output label for our disease classification), Follow-up #, Patient

ID, Patient Age, Patient Gender, View Position: X-ray orientation, image size and pixel spacing. Every patient contains a unique Patient ID and can have up to 173 follow-up X-rays. X-rays are taken in 2 different view positions, posterior anterior (PA) or anterior posterior(AP). PA X-rays are taken from the frontal chest area, and the AP positions are taken from the back. Approximately 60% of the X-ray images in the dataset are in PA view position, and 40% are in AP view position.

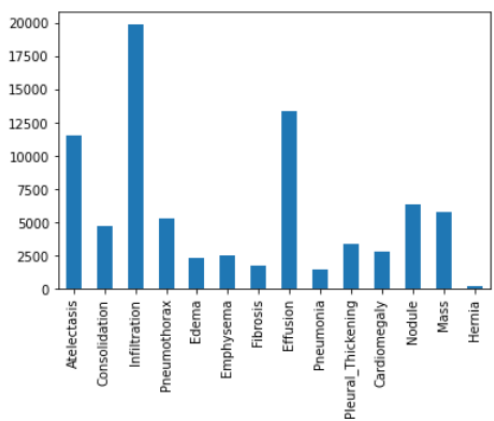

Figure 4: Frequency Distribution Excluding "No Findings"

Every image can contain upto 15 class labels within "Finding Label" . Of those 15 classes, 14 are unique diseases and 1 is "No Finding" . In the dataset, most of the X-rays are"No Findings", and the least is "Hernia".

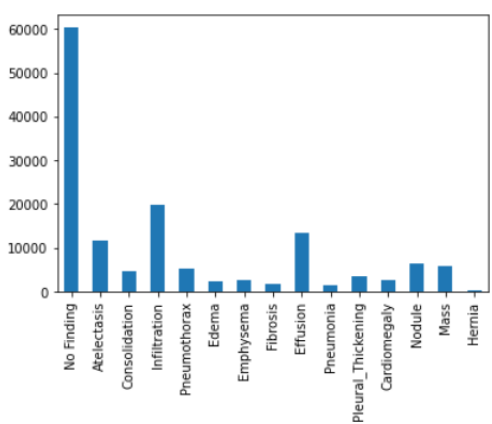

Figure 5: Frequency distribution of output labels Including"No Findings"

### 4.2.1 Dataset Preprocessing

Patient X-rays are grouped using "Patient ID" and "Follow-up #" labels. Filters are then applied on the dataset.

**Filter 1**

Patients containing less than 2 follow-ups are removed from the dataset. The filtered data then contains 9,189 patients.

**Filter 2**

Patients that contain a mix of both PA and AP view positions in their follow-up X-rays are removed. After filter 2 is applied to previously filtered data, 2,992 patients remain.

### 4.2.2 Creating Sample Sets

After applying both filters on the dataset, the sample sets were created. We define one sample set to be containing the information regarding the three followup X-rays per each patient. We group by each patient and we order it based on the followup # associated with each patient. So, the third followup contains the label for that particular sample. Each sample set consists of X-ray records of a patient with 3 consecutive follow-ups. In some cases, X-rays can belong to more than one class description and it means that X-rays can be labelled under any of these classes. In order to solve this issue, we had inspected the third follow-up for each sample set. If the 3rd follow-up contains more than 1 disease label, new sample sets are created with the same X-rays but different class description labels per each sample set. The new 3rd followup holds a new class description from the "Finding Labels". Here's an example of how we broke up such a case.

```
example:
    00000013_023.png , Infiltration|Mass|Pneumothorax    Follow-up 0, Patient 13
    00000013_024.png , Mass                              Follow-up 1, Patient 13
    00000013_025.png , Cardiomegaly|Infiltration         Follow-up 2, Patient 13

result :
    00000013_023.png , Infiltration|Mass|Pneumothorax    Follow-up 0, Patient 13, Sample # 1
    00000013_024.png , Mass                              Follow-up 1, Patient 13, Sample # 1
    00000013_025.png , Cardiomegaly                      Follow-up 2,Patient 13, Sample # 1

    00000013_023.png , Infiltration|Mass|Pneumothorax    Follow-up 0, Patient 13, Sample # 2
    00000013_024.png , Mass                              Follow-up 1,Patient 13, Sample # 2
    00000013_025.png , Infiltration                      Follow-up 2,Patient 13, Sample # 2
```

Figure 6: example

### 4.2.3 Split Sample Sets

We then created two datasets based on the view position associated with each X-ray. We call them AP and PA sample sets. These sample sets are created from separating the previously processed sample sets based on view position.

After some research on both AP and PA datasets, we have discovered that the chest X-ray dataset contains both views of AP and PA. We have decided to separate these two views of X-rays to test which view of the chest X-ray could train the neural network we built better, faster, and more accurately. We have reviewed many research papers on deep learning approaches to chest X-rays to predict thoracic diseases, but most of them did not split datasets into two and train their model separately. By splitting the dataset into two would distinguish which dataset may result in a better performance.

To observe both sides of the chest X-rays, we have separated the chest X-ray datasets into AP dataset and PA dataset to train our deep learning model and we attempted to fully understand the performance of the deep learning when we were training the model in two different datasets. This way, we can observe a clear observation of the heart and the respiratory systems of patients to easily distinguish the respiratory diseases through the multi-classification approach of the deep learning technique. Here's some statistics of how many samples finally we had and how many unique patients we are having.

|  | PA Dataset | AP Dataset |
|---|---|---|
| # of Patients | 2,552 | 440 |
| # of Samples | 8,114 | 3,905 |

### 4.2.4 Train/Test/Validation dataset split

For training purposes, we have split the dataset into AP dataset and PA dataset. For each of those datasets we split from the original dataset, we have chosen 70% as the train data, 20% as the test data, and 10% as the validation. Each of these datasets was saved as a comma-separated value (CSV). By having these datasets, we had to link the NIH X-ray images for each sample set. For the purpose of training, we have resized all images into 128 by 128 from 1024 by 1024 PNG files which captured the key regions within an X-ray image. We had rescaled the pixel values into a range of [0,1] by normalizing.

### 4.3 Basic Algorithm for the Modeling

We wanted to follow a two-part architecture for solving this problem. Since we had constructed a slightly different input setting but we had a multi classification problem. So, in order to consider this problem, we devised a simple way to combine two different sub models to solve this problem.

The first part of our architecture involves a deep convolution neural network used for feature extraction on images. We had decided to use three different deep convolution networks for image based feature extraction namely DenseNet[4], ResNet[3], and MobileNetV2[12]. We had chosen specifically: DenseNet-169, ResNet50V2, and the MobileNetV2 architectures respectively. This is basically represented as the 'CNN' network in our model diagrams. So, to this CNN model we use, we will connect 3 input X-ray images, each of size (128,128), representing the consecutive followup grayscale X-ray images of a patient. We use the same CNN model for each of our 3 input X-ray images we provide while training for one particular model. For example: we have 3 independent models: with the one single same CNN model across each image used being either ResNet, DenseNet or MobileNet for all the three images. For computational efficiency and speeding up our training task, we have decided to freeze (or not allow the model to train these parameters) the layers of these CNN models. After training from the CNN, we collect the output from the CNN model for each X-ray image and flatten each of these outputs to obtain a single dimension vector representation of each output. Then, we concatenate all these three vectors into one large vector containing the output of all the three images from the CNN model into one single vector.

In the second part of our architecture, we were able to devise two different model architectures overall. In the first architecture, we had connected our single concatenated vector output from the previous step into a Long Short Term Memory RNN layer with a conviction that there might be a time based temporal relationship across each pixel from the three images. The output of the LSTM layer is then connected to the output layer with 15 output units with a sigmoid activation for finding the probability (within a range [0,1]) for each output label and without finding the single output label.

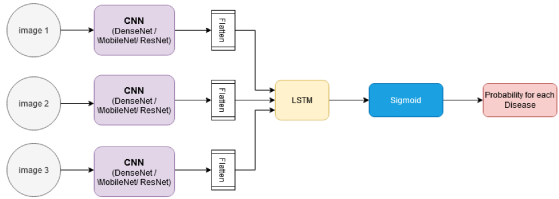

Figure 7: Model Including LSTM

The other architecture, we just never used a LSTM layer for training models. We had just connected to the output layer with 15 output units with a sigmoid activation for finding the probability (within a range [0,1]) for each output label and without finding the single output label.

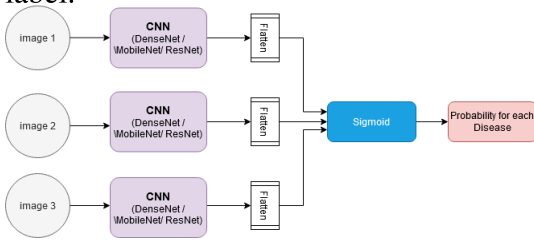

Figure 8: Model Excluding LSTM

## 5 Modeling Work/Experimental Settings

We have used DenseNet169, ResNet50-V2, and MobileNetV2 for evaluating our neural network. We have trained each of these models with and without Long Short Term Memory (LSTM) to verify our hypothesis on whether LSTM would increase the overall performance of each model we chose to use, and we attempted on 10 epochs and with a batch size of 100. We did not load any pre-trained model on any of the models and we switched 'weight' to be 'None' on each of these models when we were training. For each of the inputs to each model, we have organized the first follow-ups of all of the patients into one vector and the second follow-ups on a separate vector, and then the third follow-ups on another vector.

Each layer of each model was concatenated from the previous layers. For the LSTM layer, we have used 50 units of LSTM, with a "tanh" activation function for the first layer and used a recurrent sigmoid activation function. For the optimizer, we have chosen Adam for an efficient optimization process and the Adam optimizer has a learning rate $10^{-2}$. The dropout size was 0.2 and we had a final fully connected output layer that predicts the probability of each disease label with sigmoid activation, so we used a 15-unit Dense layer in Keras. We chose a sigmoid function because we wanted to know the probability of each disease label and we wanted to report for each disease in place of predicting one single class. This also helped us to visualize when we presented results using the ROC curve.

We have chosen binary cross-entropy for loss function and simple accuracy for determining how accurate each model is since we need to confirm whether there is a symptom or not. To determine the probability of each disease label, we need to use a binary classification loss function.

## 6 Results and Discussions

The tables below contain auc scores for the CNN models DenseNet169, MobileNetV2, and ResNet50V2. Each model was trained on either PA or AP view positions, and with LSTM either included or excluded from the model. All the models in general show a better performance on PA training data which contains more than double the number of AP samples, hence allowing better model generalizations. Moreover, PA x-rays compose 60% of the NIH dataset[15].

| Finding Labels | DenseNet169 | MobileNetV2 | ResNet50V2 |
|---|---|---|---|
| No Finding | 0.56 | 0.5 | **0.571** |
| Atelectasis | 0.44 | **0.5** | 0.47 |
| Consolidation | 0.53 | 0.5 | **0.536** |
| Infiltration | 0.58 | 0.5 | **0.591** |
| Pneumothorax | **0.56** | 0.494 | 0.512 |
| Edema | 0.52 | 0.5 | **0.622** |
| Emphysema | 0.38 | **0.5** | 0.379 |
| Fibrosis | 0.5 | **0.5** | 0.492 |
| Effusion | **0.55** | 0.5 | 0.535 |
| Pneumonia | **0.64** | 0.5 | 0.622 |
| Pleural_Thickening | **0.54** | 0.518 | 0.538 |
| Cardiomegaly | 0.52 | 0.5 | **0.56** |
| Nodule | 0.57 | 0.5 | **0.579** |
| Mass | **0.59** | 0.5 | 0.474 |
| Hernia | 0.42 | **0.5** | 0.362 |

Table 1: PA with LSTM AUC Scores

| Finding Labels | DenseNet169 | MobileNetV2 | ResNet50V2 |
|---|---|---|---|
| No Finding | 0.66 | 0.5 | **0.679** |
| Atelectasis | **0.71** | 0.5 | 0.694 |
| Consolidation | 0.63 | 0.5 | **0.67** |
| Infiltration | **0.57** | 0.5 | 0.594 |
| Pneumothorax | 0.75 | 0.5 | **0.761** |
| Edema | 0.52 | 0.5 | **0.531** |
| Emphysema | 0.6 | 0.5 | **0.67** |
| Fibrosis | 0.49 | **0.5** | 0.47 |
| Effusion | 0.7 | 0.5 | **0.71** |
| Pneumonia | **0.7** | 0.5 | 0.664 |
| Pleural_Thickening | 0.59 | 0.5 | **0.61** |
| Cardiomegaly | 0.61 | 0.5 | **0.699** |
| Nodule | **0.56** | 0.5 | 0.553 |
| Mass | 0.63 | 0.5 | **0.643** |
| Hernia | 0.71 | 0.5 | **0.811** |

Table 3: PA without LSTM AUC Scores

| Finding Labels | DenseNet169 | MobileNetV2 | ResNet50V2 |
|---|---|---|---|
| No Finding | 0.502 | 0.5 | **0.562** |
| Atelectasis | 0.444 | **0.5** | 0.447 |
| Consolidation | **0.544** | 0.5 | 0.525 |
| Infiltration | **0.529** | 0.5 | 0.465 |
| Pneumothorax | 0.378 | 0.5 | **0.599** |
| Edema | 0.509 | 0.5 | **0.527** |
| Emphysema | **0.658** | 0.5 | 0.633 |
| Fibrosis | 0.554 | 0.5 | **0.749** |
| Effusion | 0.522 | 0.492 | **0.567** |
| Pneumonia | 0.431 | 0.468 | **0.569** |
| Pleural_Thickening | **0.625** | 0.5 | 0.435 |
| Cardiomegaly | **0.529** | 0.5 | 0.423 |
| Nodule | **0.603** | 0.5 | 0.55 |
| Mass | 0.288 | **0.498** | 0.278 |
| Hernia | - | - | - |

Table 2: AP with LSTM AUC Scores

| Finding Labels | DenseNet169 | MobileNetV2 | ResNet50V2 |
|---|---|---|---|
| No Finding | **0.691** | 0.5 | 0.684 |
| Atelectasis | **0.6** | 0.5 | 0.567 |
| Consolidation | 0.585 | 0.5 | **0.618** |
| Infiltration | **0.555** | 0.5 | 0.509 |
| Pneumothorax | **0.824** | 0.5 | 0.805 |
| Edema | **0.609** | 0.5 | 0.599 |
| Emphysema | 0.812 | 0.5 | **0.834** |
| Fibrosis | **0.743** | 0.5 | 0.592 |
| Effusion | **0.572** | 0.5 | 0.541 |
| Pneumonia | 0.565 | 0.5 | **0.62** |
| Pleural_Thickening | 0.623 | 0.5 | **0.623** |
| Cardiomegaly | 0.594 | 0.5 | **0.605** |
| Nodule | **0.697** | 0.5 | 0.611 |
| Mass | 0.51 | 0.5 | **0.682** |
| Hernia | - | - | - |

Table 4: AP without LSTM AUC Scores

There is no observable time relationship between 3 images' pixels in a set after observing auc scores between CNN models because the models perform better when LSTM layer is not included in the architecture. So, in terms of the three internal Convolutional neural network models used, we constantly observed that in most of the 4 cases: (PA/AP with/without LSTMs) ResNet50V2 does generally better than the DenseNet169 models while the MobileNetv2 performs the worst in most of the cases. The main reason we can possibly attribute for such behavior is probably due to the architecture of each of these deep learning frameworks. While ResNet has 50 hidden layers in its deep learning architecture, the MobileNetV2 possesses a fully convolution layer with 32 filters, and then 19 residual bottleneck layers.

So, probably for understanding the features within 3 followup X-rays per sample, we suspect deeper learning framework of CNNs perform better feature extraction techniques rather than a lighter version. In comparison with DenseNet, surprisingly again ResNet does mostly better (except in case of AP without LSTM where mostly DenseNet does better). We again attempt to explain this behavior by observing the number of parameters associated to work on the images. To be precise, we find that the number of parameters associated with the ResNet50 model is 23,558,528 parameters, while DenseNet169 model has 12,636,608 parameters and MobileNetV2 model has 2,257,408 parameters for the CNN itself. So, it can be understood that the CNN model with much more parameters tends to learn better attributing the fact that there are probably more convolutional layers which can possibly extract the key regions within an X-ray image and hence improve the disease classification.

| Labels | 1 Img (PA) | 3 Img (PA) | 1 Img (AP) | 3 Img (AP) |
|---|---|---|---|---|
| Atelectasis | 0.585 | **0.71** | 0.562 | **0.6** |
| Consolidation | 0.558 | **0.63** | 0.585 | **0.585** |
| Infiltration | 0.473 | **0.57** | 0.518 | **0.555** |
| Pneumothorax | 0.564 | **0.75** | 0.792 | **0.824** |
| Edema | 0.334 | **0.52** | **0.667** | 0.609 |
| Emphysema | 0.527 | **0.6** | 0.576 | **0.812** |
| Fibrosis | 0.334 | **0.49** | 0.619 | **0.743** |
| Effusion | **0.752** | 0.7 | **0.63** | 0.572 |
| Pneumonia | 0.523 | **0.7** | 0.362 | **0.565** |
| Pleural_Thickening | **0.594** | 0.59 | 0.558 | **0.623** |
| Cardiomegaly | **0.682** | 0.61 | 0.514 | **0.594** |
| Nodule | 0.448 | **0.56** | 0.504 | **0.697** |
| Mass | 0.569 | **0.63** | **0.628** | 0.51 |
| Hernia | - | **0.71** | - | - |

Table 5: Single vs 3 images as input: Densenet AUC Scores Without LSTM

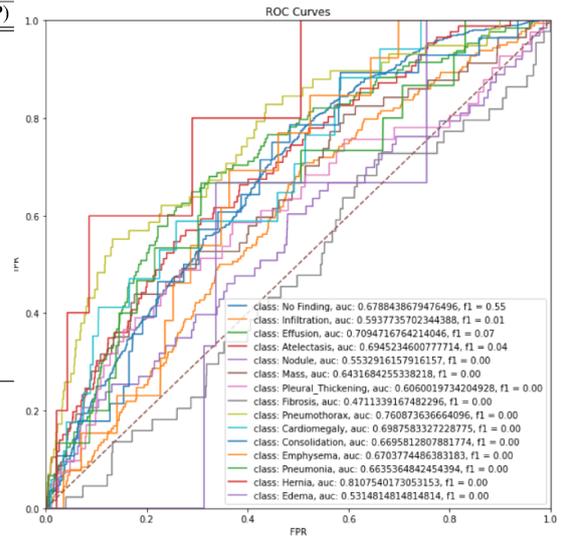

Figure 9: ResNet PA ROC

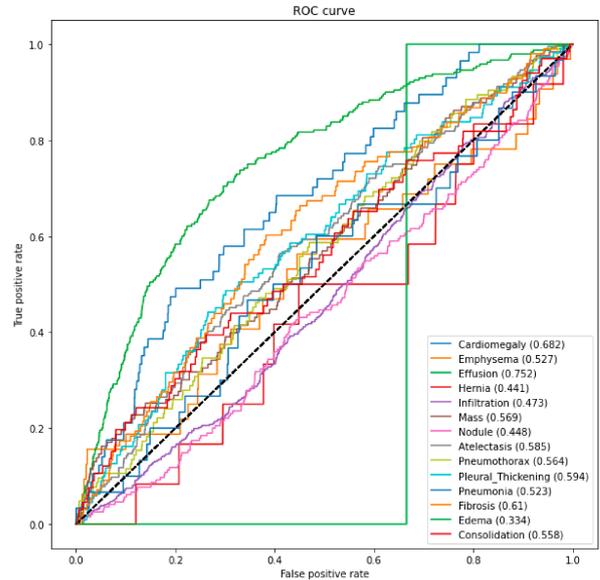

Figure 10: Single Image PA ROC

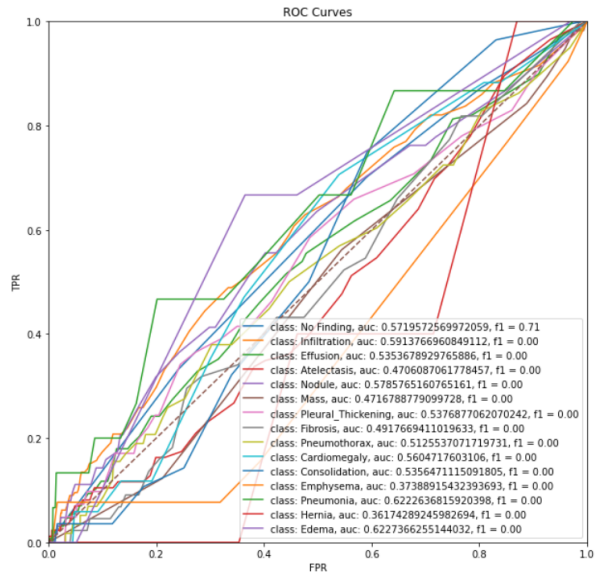

In case of single images as input and three X-ray images as input, we observe that the three image input performs better than a single image baseline model. This is a clear proof of our hypothesis that 3 followup X-rays as input would perform much better than a single image based disease classification. We again attribute this fact with the reason that three images provide much more information to analyze the diseases in comparison to a single image as an input and hence CNNs probably are able to capture a better trend across the 3 followup images. We are able to arrive at a final conclusion that more than one image input can perform better than a single image input given the same CNN model used for feature extraction.

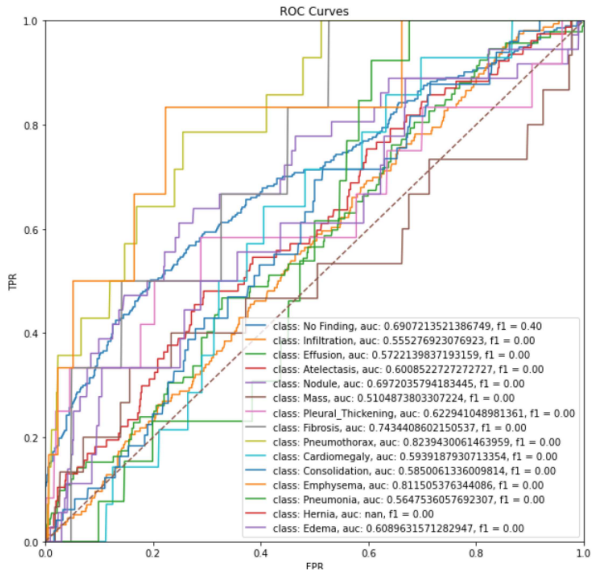

Figure 11: ResNet LSTM PA ROC

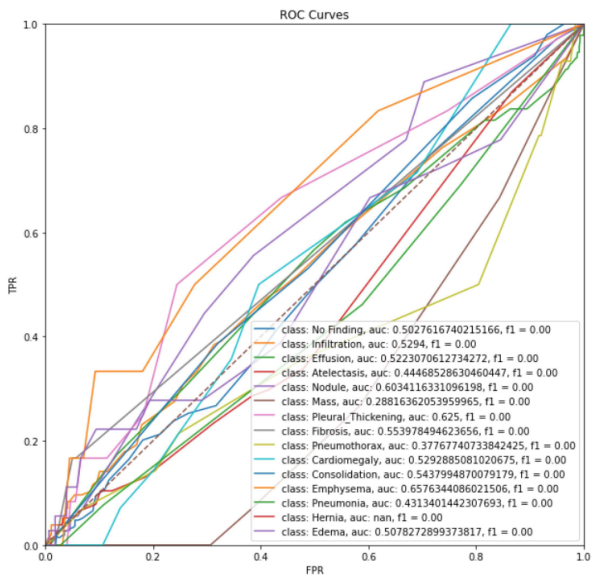

Figure 12: DenseNet AP ROC

Figure 13: DenseNet LSTM AP ROC

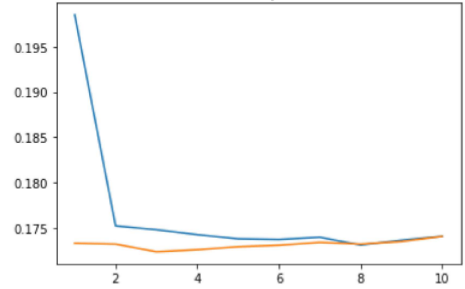

Figure 14: DenseNet LSTM PA Loss

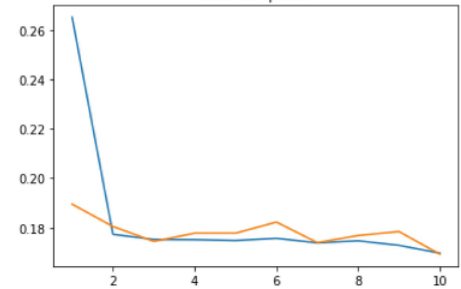

Figure 15: DenseNet PA Loss

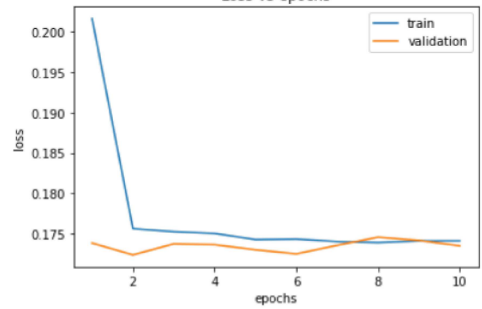

Figure 16: ResNet LSTM PA Loss

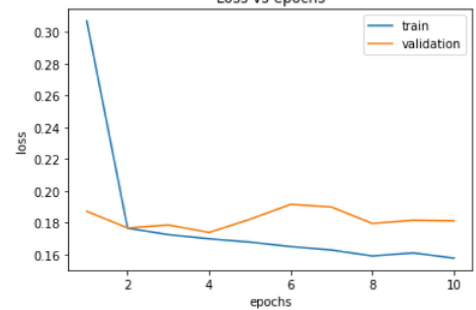

Figure 17: ResNet PA Loss

# 7 Conclusions

We have established a different setting to a traditional disease classification problem. We had used 3 input X-rays which are the consecutive previous X-ray images of every patient based on their previous history. We had split into different datasets based on their view position: AP view position and PA view position. Then, we have created and validated two different deep neural network architectures (with/without LSTM) using AP and PA dataset with different CNN architectures such as DenseNet169, MobileNetV2, or ResNet50V2 for feature extraction. According to our results, architecture without LSTM outperforms architecture with LSTM for using the same setting and internal layer. Among the CNN internal layers, ResNet without LSTM performs the best to predict the disease label based on the differences occurring across given follow-up X-rays according to ROC-AUC scores. Among the CNN layers we experimented, we have discovered and proven that ResNet50V2 without LSTM performs the best to predict the disease labels. We have also proven our hypothesis that having three sequences of chest X-ray of one patient can increase the accuracy of predicting the next disease better than having only one sequence of chest x-ray of a patient. We have collected and organized ROC-AUC analysis for demonstrating the differences across the algorithms we chose, and we have shown overall good performance in the research.

For future analysis, we plan to use GradCAM technique on our best models for identification of the regions on custom X-rays which may be crucial to classify the disease label. Also, using three followup images was more of a convenient choice in comparison to using a single image and we could have experimented with more follow up X-ray images for a stronger relationship across results.

# 8 Contributions

## 8.1 Seongwoo Choi (shjchoi@ucdavis.edu):

- Wrote and edited the report and chose the topic on AI in Medicine using chest X-ray.
- Coded the single baseline, dataset preprocessing, and data visualization.
- Choi also uploaded the entire dataset from the National Institutes of Health on the shared drive and his teammates could utilize the dataset with ease.
- Reviewed research papers that related to the project.
- Trained DenseNet on AP with and without LSTM and recoreded the results.
- Maintained and debugged codes with teammates and managed GitHub repository.

## 8.2 Shikun Huang (kunhuang@ucdavis.edu)

- Wrote code to analyze NIH dataset details and data distribution.
- Coded Python code for dataset preprocessing with filters, creating sample sets from preprocessed dataset, and sample set X-ray image extraction.
- Trained MobileNetV2 and ResNet50V2 on AP view position models; coded and research on ROC-AUC curve and F1-score implementation for multi-classification on python
- Managed and cleaned github repository; Wrote README for github repository
- Reviewed research papers for related work
- Edited the report

## 8.3 David Haddad (davhaddad@ucdavis.edu)

- Wrote code to Analyze NIH dataset details. Displayed bar charts and pi charts
- Wrote Python code for dataset preprocessing with filters, creating sample sets from preprocessed dataset, and sample set X-ray image extraction.
- Trained MobileNetV2 and ResNet50V2 PA view position models

- Reviewed research papers for related work
- Worked on model diagram and overall pipeline for report
- Worked on paper dataset, methodology and result tables and plots for report

### 8.4 Sairamvinay Vijayaraghavan (saivijay@ucdavis.edu)

- Devised the basic idea for this project: the 3 image input of X-rays and deep learning CNN models as a sequence
- Development of ideas along with the team: worked on devising the solutions and the hypothesis for this paper
- Wrote Python Code for the basic script to adapt for creating other models for the single image prediction for the disease classification and the triple image disease prediction.
- Trained the single image PA baseline model ; Also the DenseNet PA models with/without LSTMs
- Edited and fine tuned the report sections for the term paper.
- Reviewed research papers for related work.

# References

[1] Ivo M Baltruschat, Hannes Nickisch, Michael Grass, Tobias Knopp, and Axel Saalbach. Comparison of deep learning approaches for multi-label chest x-ray classification. *Scientific reports*, 9(1):1–10, 2019.

[2] Sebastian Guendel, Florin C Ghesu, Sasa Grbic, Eli Gibson, Bogdan Georgescu, Andreas Maier, and Dorin Comaniciu. Multi-task learning for chest x-ray abnormality classification on noisy labels. *arXiv preprint arXiv:1905.06362*, 2019.

[3] Kaiming He, Xiangyu Zhang, Shaoqing Ren, and Jian Sun. Identity mappings in deep residual networks. volume 9908, pages 630–645, 10 2016.

[4] G. Huang, Z. Liu, L. Van Der Maaten, and K. Q. Weinberger. Densely connected convolutional networks. In *2017 IEEE Conference on Computer Vision and Pattern Recognition (CVPR)*, pages 2261–2269, 2017.

[5] A Mueen, Sapiyan Baba, and Roziati Zainuddin. Multilevel feature extraction and x-ray image classification. *Journal of Applied Sciences*, 7(8):1224–1229, 2007.

[6] Andrew Ng, Pranav Rajpurkar, Amirhossein Kiani, and Eddy Shyu. Ai in medicine - deeplearning.ai, 2020.

[7] Harsh Panwar, PK Gupta, Mohammad Khubeb Siddiqui, Ruben Morales-Menendez, Prakhar Bhardwaj, and Vaishnavi Singh. A deep learning and grad-cam based color visualization approach for fast detection of covid-19 cases using chest x-ray and ct-scan images. *Chaos, Solitons & Fractals*, 140:110190, 2020.

[8] Dr Graham Lloyd-Jones BA MBBS MRCP FRCR Consultant Radiologist. Chest x-ray quality projection.

[9] Pranav Rajpurkar, Jeremy Irvin, Robyn L Ball, Kaylie Zhu, Brandon Yang, Hershel Mehta, Tony Duan, Daisy Ding, Aarti Bagul, Curtis P Langlotz, et al. Deep learning for chest radiograph diagnosis: A retrospective comparison of the chexnext algorithm to practicing radiologists. *PLoS medicine*, 15(11):e1002686, 2018.

[10] Pranav Rajpurkar, Jeremy Irvin, Kaylie Zhu, Brandon Yang, Hershel Mehta, Tony Duan, Daisy Ding, Aarti Bagul, Curtis Langlotz, Katie Shpanskaya, et al. Chexnet: Radiologist-level pneumonia detection on chest x-rays with deep learning. *arXiv preprint arXiv:1711.05225*, 2017.

[11] H Roopa and T Asha. Feature extraction of chest x-ray images and analysis using pca and kpca. *International Journal of Electrical and Computer Engineering*, 8(5):3392, 2018.

[12] M. Sandler, A. Howard, M. Zhu, A. Zhmoginov, and L. Chen. Mobilenetv2: Inverted residuals and linear bottlenecks. In *2018 IEEE/CVF Conference on Computer Vision and Pattern Recognition*, pages 4510–4520, 2018.

[13] Ramprasaath R Selvaraju, Michael Cogswell, Abhishek Das, Ramakrishna Vedantam, Devi Parikh, and Dhruv Batra. Grad-cam: Visual explanations from deep networks via gradient-based localization. In *Proceedings of the IEEE international conference on computer vision*, pages 618–626, 2017.

[14] Hongyu Wang and Yong Xia. Chestnet: A deep neural network for classification of thoracic diseases on chest radiography. 07 2018.

[15] Xiaosong Wang, Yifan Peng, Le Lu, Zhiyong Lu, Mohammadhadi Bagheri, and Ronald M Summers. Chestx-ray8: Hospital-scale chest x-ray database and benchmarks on weakly-supervised classification and localization of common thorax diseases. In *Proceedings of the IEEE conference on computer vision and pattern recognition*, pages 2097–2106, 2017.

[16] Fang Yang, Murat Hamit, Chuan B Yan, Juan Yao, Abdugheni Kutluk, Xi M Kong, and Sui X Zhang. Feature extraction and classification on esophageal x-ray images of xinjiang kazak nationality. *Journal of healthcare engineering*, 2017, 2017.